\documentclass[11pt,a4paper]{article}
\usepackage[hyperref]{naaclhlt2018}
\usepackage{hyperref}
\usepackage{times}
\usepackage{latexsym}
\usepackage{url}
\usepackage{amsmath}
\usepackage{hhline}
\usepackage[T1]{fontenc}
\usepackage{graphicx}
\usepackage{eurosym}

\usepackage{kantlipsum}

\usepackage{tabularx}
\graphicspath{ {images/} }
\usepackage{listings}
\usepackage{stfloats}
\usepackage{csquotes}
\usepackage{array}
\usepackage{booktabs}
\usepackage[super]{nth}
\usepackage[inline,shortlabels]{enumitem}
\usepackage{subfig}
\usepackage{amsmath}
\usepackage{multirow}
\usepackage{enumitem}

\usepackage{svg}
\usepackage{float}
\usepackage{makecell}

\usepackage[scientific-notation=true]{siunitx}
\usepackage{marginnote}
\usepackage{amssymb}
\usepackage{pifont}
\newcommand{\cmark}{\ding{51}}%
\usepackage[textsize=tiny]{todonotes}
\usepackage{mdframed}
\usepackage{dirtytalk}
\usepackage{xprintlen}
\DeclareMathOperator*{\argmax}{arg\,max}
\usepackage{todonotes}
\aclfinalcopy 
 
\title{NTUA-SLP at IEST 2018: Ensemble of Neural Transfer Methods for Implicit Emotion Classification}

\author{ 
	Alexandra Chronopoulou$^{1}$\thanks{*These authors contributed equally to this work.}, Aikaterini Margatina$^{1 *}$  \\ 
	{\bf Christos Baziotis$^{1,2}$, Alexandros Potamianos$^{1,3}$} \\\\
	$^1$School of ECE, National Technical University of Athens, Athens, Greece \\
	$^2$ Department of Informatics, Athens University of Economics and Business, Athens, Greece \\
	$^3$ Signal Analysis and Interpretation Laboratory (SAIL), USC, Los Angeles, USA\\  
    {\tt}\\
	{\tt el12068@central.ntua.gr, el12108@central.ntua.gr} \\
    {\tt cbaziotis@mail.ntua.gr, potam@central.ntua.gr}\\
}

\date{2018}

\begin{document}

\maketitle

\begin{abstract}
In this paper we present our approach to tackle the Implicit Emotion Shared Task (IEST) organized as part of WASSA 2018 at EMNLP 2018.
Given a tweet, from which a certain word has been removed, we are asked to predict the emotion of the missing word.
In this work, we experiment with neural Transfer Learning (TL) methods. 
Our models are based on LSTM networks, augmented with a self-attention mechanism.
We use the weights of various pretrained models,  for initializing specific layers of our networks.
We leverage a big collection of unlabeled Twitter messages, for pretraining word2vec word embeddings and a set of diverse language models. Moreover, we utilize a sentiment analysis dataset for pretraining a model, which encodes emotion related information.
The submitted model consists of an ensemble of the aforementioned TL models.
Our team ranked \nth{3} out of 30 participants, achieving an $F_{1}$ score of 0.703.

\end{abstract}

\section{Introduction}
Social media, especially micro-blogging services like Twitter, have attracted lots of attention from the NLP community.  
The language used is constantly evolving by incorporating new syntactic and semantic constructs, such as emojis or hashtags, abbreviations and slang, making natural language processing in this domain even more demanding. 
Moreover, the analysis of such content leverages the high availability of datasets offered from Twitter, satisfying the need for large amounts of data for training.
    
    Emotion recognition is particularly interesting in social media, as it has useful applications in numerous tasks, such as public opinion detection about political  tendencies~\cite{pla2014political,tumasjan2010predicting,li2014text}, stock market monitoring~\cite{si2013exploiting,bollen2011twitter}, tracking product perception~\cite{chamlertwat2012discovering}, even detection of suicide-related communication \cite{burnap2015machine}.
    
    	In the past,  emotion analysis, like most NLP tasks, was tackled by traditional methods that included hand-crafted features or features from sentiment lexicons~\cite{nielsen2011new,mohammad2010emotions,mohammad2013crowdsourcing,go2009twitter} which were fed to classifiers such as Naive Bayes and SVMs~\cite{bollen2011modeling,mohammad2013a,kiritchenko2014}. However, deep neural networks achieve increased performance compared to traditional methods, due to their ability to learn more abstract features from large amounts of data, producing state-of-the-art results in emotion recognition and sentiment analysis~\cite{deriu2016swisscheese,goel2017prayas, baziotis2017datastories}.
    
    In this paper, we present our work submitted to the WASSA 2018 IEST~\cite{Klinger2018x}. In the given task, the word that triggers emotion is removed from each tweet and is replaced by the token \texttt{[\#TARGETWORD\#]}. The objective is to predict its emotion category among 6 classes: \textit{anger, disgust, fear, joy, sadness} and \textit{surprise}. Our proposed model employs 3 different TL schemes of pretrained models: word embeddings, a sentiment model and language models.




\begin{figure*}[t]
	\captionsetup{farskip=0pt} 
	\centering
	\includegraphics[width=.8\textwidth, page=2]{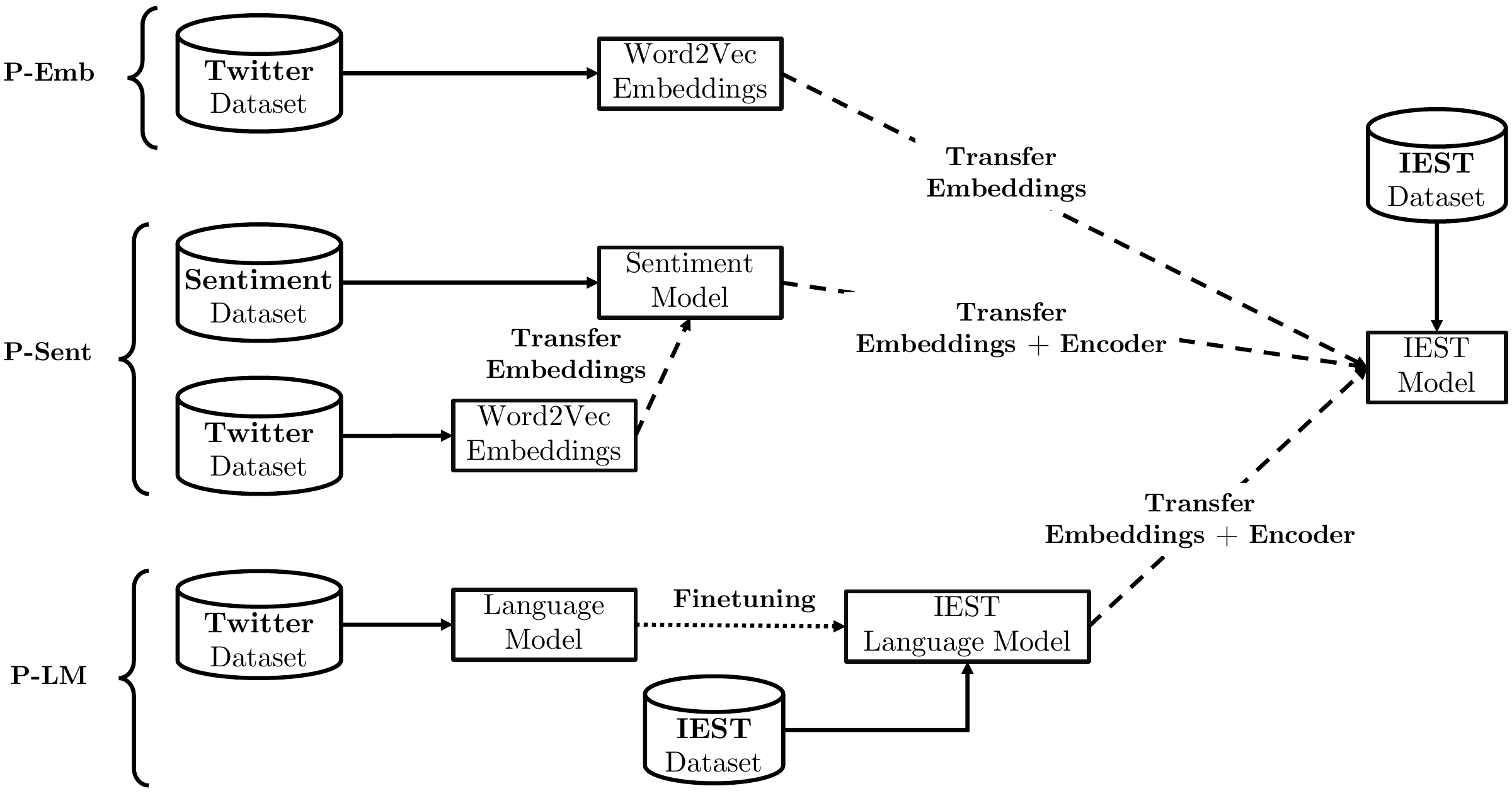}
	\caption{High-level overview of our TL approaches.}
	\label{fig:overview}
\end{figure*}

\section{Overview}

Our approach is composed of the following three steps:
(1) \textit{pretraining}, in which we train word2vec word embeddings (P-Emb), a sentiment model (P-Sent) and Twitter-specific language models (P-LM),
(2) \textit{transfer learning}, in which we transfer the weights of the aforementioned models to specific layers of our IEST classifier and
(3) \textit{ensembling}, in which we combine the predictions of each TL model.
Figure \ref{fig:overview} depicts a high-level overview of our approach.

\subsection{Data}
\label{sec:data}
Apart from the IEST dataset, we employ a SemEval dataset for sentiment classification and other manually-collected unlabeled corpora for our language models.

\noindent\textbf{Unlabeled Twitter Corpora.} We collected a dataset of 550 million archived English Twitter messages, from 2014 to  2017. This dataset is used for calculating word statistics for our text preprocessing pipeline and training our \textit{word2vec} word embeddings presented in Sec. \ref{sec:system1}.

For training our language models, described in Sec. \ref{sec:system3}, we sampled three subsets of this corpus. The first consists of 2M tweets, all of which contain emotion words. To create the dataset, we selected tweets that included one of the six emotion classes of our task (\textit{anger, disgust, fear, joy, sadness} and \textit{surprise}) or synonyms. We ensured that this dataset is balanced by concatenating approximately 350K tweets from each category. The second chunk has 5M tweets, randomly selected from the initial 550M corpus. We aimed to create a general sub-corpus, so as to focus on the structural relationships of words, instead of their emotional content. The third chunk is composed of the two aforementioned corpora. We concatenated the 2M emotion dataset with 2M generic tweets, creating a final 4M dataset. We denote the three corpora as \textit{EmoCorpus} (2M), \textit{EmoCorpus+} (4M) and \textit{GenCorpus} (5M).

\noindent\textbf{Sentiment Analysis Dataset}.  We use the dataset of SemEval17 Task4A (Sent17)~\cite{rosenthal2017semeval} for training our sentiment classifier as described in Sec. \ref{sec:system2}. The dataset consists of Twitter messages annotated with their sentiment polarity (\textit{positive, negative, neutral}). The training set contains 56K tweets and the validation set 6K tweets.

\subsection{Preprocessing} \noindent To preprocess the tweets, we use \textit{Ekphrasis}~\citep{baziotis2017datastories}, a tool geared towards text from social networks, such as Twitter and Facebook. \textit{Ekphrasis} performs Twitter-specific tokenization, spell correction, word normalization, segmentation (for splitting hashtags) and annotation.

\subsection{Word Embeddings}
Word embeddings are dense vector representations of words which capture  semantic and syntactic information. For this reason, we employ the \textit{word2vec}~\cite{mikolov2013} algorithm to train our word vectors, as described in Sec. \ref{sec:system1}.

\subsection{Transfer Learning}
Transfer Learning (TL) uses knowledge from a learned task so as to improve the performance of a related task by reducing the required training data~\cite{torrey2010transfer, pan2010survey}.  In computer vision, transfer learning is employed in order to overcome the deficit of training samples for some categories by adapting classifiers trained for other categories~\cite{oquab2014learning}. With the power of deep supervised learning, learned knowledge can even be transferred to a totally different task (i.e. \textit{ImageNet}~\cite{krizhevsky2012imagenet}). 

Following this logic, TL methods have also been applied to NLP. Pretrained word vectors~\cite{mikolov2013,pennington2014} have become standard components of most architectures.
Recently, approaches that leverage pretrained language models have emerged, which learn the compositionality of language, capture long-term dependencies and context-dependent features. For instance, ELMo contextual word representations~\cite{peters2018deep} and ULMFiT ~\cite{DBLP:journals/corr/abs-1801-06146} achieve state-of-the-art results on a wide variety of NLP tasks. Our work is mainly inspired by ULMFiT, which we extend to the Twitter domain.


\subsection{Ensembling} \label{sec::ensemble}
We combine the predictions of our 3 TL schemes with the intent of increasing the generalization ability of the final classifier. To this end, we employ a pretrained word embeddings approach, as well as a pretrained sentiment model and a pretrained LM. We use two ensemble schemes, namely unweighted average and majority voting.

\noindent\textbf{Unweighted Average (UA)}.
In this approach, the final prediction is estimated from the unweighted average of the posterior probabilities for all different models.  Formally, the final prediction $p$ for a training instance is estimated by:
\begin{align}
p &= \argmax_{c}\frac{1}{C} \sum_{i=1}^{M} \vec{p_i}
\label{eq:att_ei}, \space \quad p_i \in {\rm I\!R}^C
\end{align}
where $C$ is the number of classes, $M$ is the number of different models, $c \in \{1,...,C\}$ denotes one class and $\vec{p_i}$ is the probability vector calculated by model $i \in \{1,...,M\}$ using softmax function. \par
\noindent\textbf{Majority Voting (MV)}.
Majority voting approach counts the votes of all different models and chooses the class with most votes. Compared to UA, MV is affected less by single-network decisions. However, this schema does not consider any information derived from the minority models. Formally, for a task with $C$ classes and $M$ different models, the prediction for a specific instance is estimated as follows: 
\begin{align}
\label{eqn:eqlabel}
\begin{split}
v_c &=  \sum_{i=1}^{M} F_i(c) \\
p &= \argmax_{c \in \{1,...,C\}} v_c 
\end{split}
\end{align}
where $v_c$ denotes the votes for class $c$ from all different models, $F_i$ is the decision of the $i^{th}$ model, which is either 1 or 0 with respect to whether the model has classified the instance in class $c$ or not and $p$ is the final prediction.
\begin{figure*}[!ht]
	\captionsetup{farskip=0pt} 
	\centering
	\includegraphics[width=0.75\textwidth]{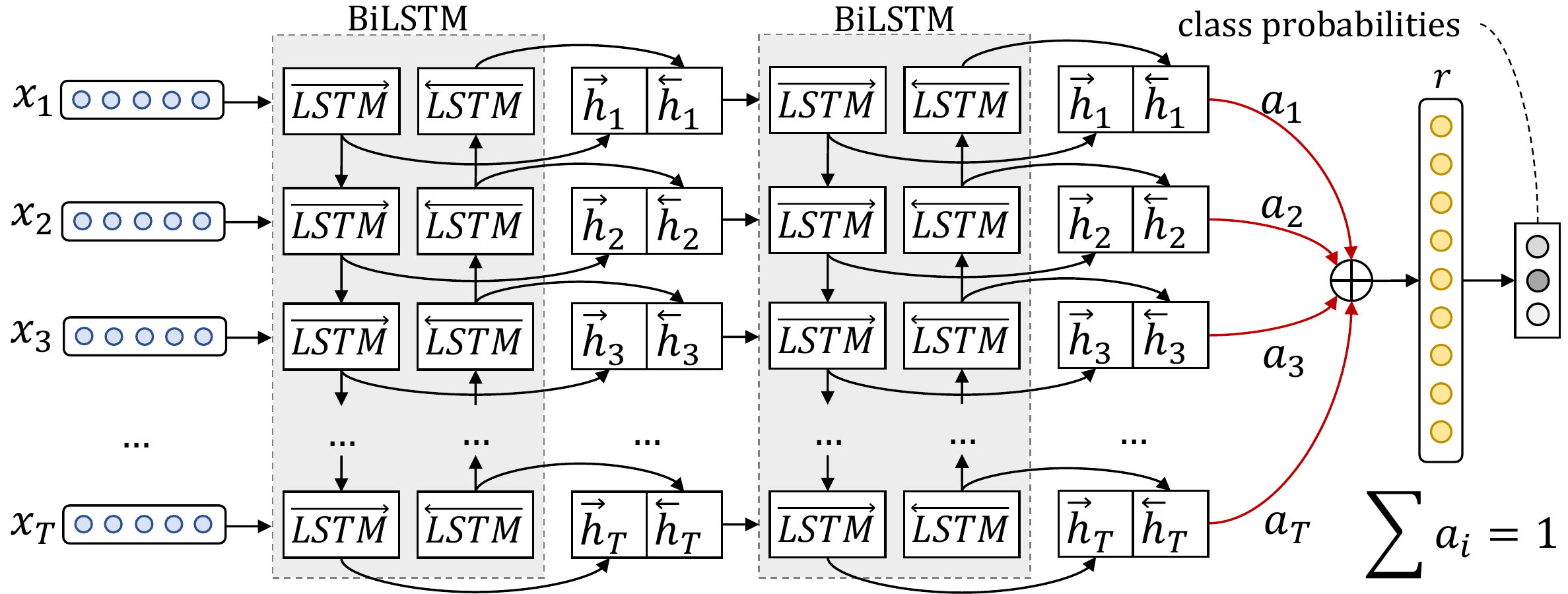}
	\caption{The proposed model, composed of a 2-layer bi-LSTM with a deep self-attention mechanism. When the model is initialized with pretrained LMs, we use unidirectional LSTM instead of bidirectional.}
	\label{fig:nn1}
\end{figure*}

\section{Network Architecture} \label{section3}
All of our TL schemes share the same architecture: 
A 2-layer LSTM with a self-attention mechanism. It is shown in Figure \ref{fig:nn1}.

\noindent\textbf{Embedding Layer}. The input to the network is a Twitter message, treated as a sequence of words. We use an embedding layer to project the words $w_1,w_2,...,w_N$ to a low-dimensional vector space $ R^W$, where $W$ is the size of the embedding layer and $N$ the number of words in a tweet. 

\noindent\textbf{LSTM Layer}. An LSTM takes as input a sequence of word embeddings and produces word annotations $h_1,h_2,...,h_N$, where $h_i $ is the hidden state at time-step $i$, summarizing  all the information of the sentence up to $w_i$. 
We use bidirectional LSTM to get word annotations that summarize the information from both directions. A bi-LSTM consists of a forward $ \overrightarrow{f} $  that parses the sentence from $w_1$ to $w_N$ and a backward $ \overleftarrow{f} $ that parses it from $w_N$ to $w_1$. We obtain the final annotation for each word $h_i$, by concatenating the annotations from both directions,
$h_i = \overrightarrow{h_i} \parallel \overleftarrow{h_i}, \quad h_i \in R^{2L}$,
where $ \parallel $ denotes the concatenation operation and $L$ the size of each LSTM. 
When the network is initialized with pretrained LMs, we employ unidirectional instead of bi-LSTMs.

\noindent\textbf{Attention Layer}. 
To amplify the contribution of the most informative words, we augment our LSTM with an attention mechanism, which assigns a weight $a_i$ to each word annotation $h_i$. We compute the fixed representation $r$ of the whole input message, as the weighted sum of all the word annotations.
\noindent
\begin{align}
e_i &= tanh(W_h h_i + b_h)\label{eq:att_ei}, \quad e_i \in [-1,1]\\
a_i &= \dfrac{exp(e_i)}{\sum_{t=1}^{T} exp(e_t)}\label{eq:att_ai}, \quad \sum_{i=1}^{T} a_i = 1 
\end{align}
\begin{align}
r &= \sum_{i=1}^{T} a_ih_i \label{eq:att_r}, \quad r \in R^{2L}
\end{align}
where $ W_h $ and $ b_h $ are the attention layer's weights.
\noindent\textbf{Output Layer}.
We use the representation $r$ as feature vector for classification and we feed it to a fully-connected softmax layer with $L$ neurons, which outputs a probability distribution over all classes $p_c$ as described in Eq.~\ref{e:outlay}:

\begin{equation}
\label{e:outlay}
p_c = \frac{e^{Wr + b}}{\sum_{i \in [1,L]}(e^{W_i r + b_i})}
\end{equation}
where $W$ and $b$ are the layer's weights and biases.

\section{Transfer Learning Approaches}
\label{sec:TL}

\subsection{Pretrained Word Embeddings (\textit{P-Emb})}
\label{sec:system1}
In the first approach, we train \textit{word2vec} word embeddings with which we initialize the embedding layer of our network. The weights of the embedding layer remain frozen during training. 
The \textit{word2vec} word embeddings are trained on the 550M Twitter corpus (Sec. \ref{sec:data}), with negative sampling of 5 and minimum word count of 20, using Gensim's~\cite{rehurek_lrec} implementation. The resulting vocabulary contains $800,000$ words. 

\subsection{Pretrained Sentiment Model (\textit{P-Sent})}
\label{sec:system2}
In the second approach, we first train a sentiment analysis model on the Sent17 dataset, using the architecture described in Sec. \ref{section3}. The embedding layer of the network is initialized with our pretrained word embeddings. Then, we fine-tune the network on the IEST task, by replacing its last layer with a task-specific layer. 

\subsection{Pretrained Language Model (\textit{P-LM})}
\label{sec:system3}
The third approach consists of the following steps: (1) we first train a language model on a generic Twitter corpus, (2) we fine-tune the LM on the task at hand and finally, (3) we transfer the embedding and RNN layers of the LM, we add attention and output layers and fine-tune the model on the target task.

\noindent\textbf{LM Pretraining}.
We collect three Twitter datasets as described in Sec. \ref{sec:data} and for each one we train an LM. In each dataset we use the 50,000 most frequent words as our vocabulary. Since the literature concerning LM transfer learning is limited, especially in the Twitter domain, we aim to explore the desired characteristics of the pretrained LM. To this end, our contribution in this research area lies in experimenting with a task-relevant corpus (EmoCorpus), a generic one (GenCorpus) and a mixture of both (EmoCorpus+). 



\noindent\textbf{LM Fine-tuning}. 
This step is crucial since, albeit the diversity of the general-domain data used for pretraining, the data of the target task will likely have a different distribution. 

	We thus fine-tune the three pretrained LMs on the IEST dataset, employing two approaches. The first is simple fine-tuning, according to which all layers of the model are trained simultaneously. The second one is a simplified yet similar approach to \textit{gradual unfreezing}, proposed in~\cite{DBLP:journals/corr/abs-1801-06146}, which we denote as \textit{Simplified Gradual Unfreezing} (SGU). According to this method, after we have transfered the pretrained embedding and LSTM weights, we let only the output layer fine-tune for $n-1$ epochs. At the $n^{th}$ epoch, we unfreeze both LSTM layers. We let the model fine-tune, until epoch $k-1$. Finally, at epoch $k$, we also unfreeze the embedding layer and let the network train until convergence. In other words, we experiment with pairs of numbers of epochs, \{n, k\}, where $n$ denotes the epoch when we unfreeze the LSTM layers and $k$ the epoch when we unfreeze the embedding layer. Naive fine-tuning poses the risk of catastrophic forgetting, or else abruptly losing the knowledge of a previously learnt task, as information relevant to the current task is incorporated. Therefore, to prevent this from happening, we unfreeze the model starting from the last layer, which is task-specific, and after some epochs we progressively unfreeze the next, more general layers, until all layers are unfrozen. 



\noindent\textbf{ LM Transfer}. This is the final step of our TL approach. We now have several LMs from the second step of the procedure. We transfer their embedding and RNN weights to a final target classifier. We again experiment with both simple and more sophisticated fine-tuning techniques, to find out which one is more helpful to this task. 

Furthermore, we introduce the \textit{concatenation method} which was inspired by the correlation of language modeling and the task at hand. We use pretrained LMs to leverage the fact that the task is basically a cloze test. In an LM, the probability of occurrence of each word, is conditioned on the preceding context, $P(w_t|w_1, \ldots, w_{t-1})$. In RNN-based LMs, this probability is encoded in the hidden state of the RNN, $P(w_t|h_{t-1})$. 
To this end, we concatenate the hidden state of the LSTM, right before the missing word, $h_{implicit}$, to the output of the self-attention mechanism, $r$: 
\begin{equation}
r' = r \parallel h_{implicit} , \quad h_i \in R^{2L}
\end{equation}
where $L$ is the size of each LSTM, and then feed it to the output linear layer. This way, we preserve the information which implicitly encodes the probability of the missing word.

\section{Experiments and Results}
\subsection{Experimental Setup}\label{sec:setup}
\noindent\textbf{Training}\label{sec:setup}.
We use Adam algorithm~\cite{kingma2014} to optimize our networks, with mini-batches of size 64 and clip the norm of the gradients~\cite{pascanu2013a} at 0.5, as an extra safety measure against exploding gradients. We also used PyTorch \cite{paszke2017automatic} and Scikit-learn \cite{pedregosa2011}.

\noindent\textbf{Hyperparameters}\label{sec:hparams}. 
For all our models, we employ the same 2-layer attention-based LSTM architecture (Sec. \ref{section3}). All the hyperparameters used are shown in Table \ref{table:hparams}.

\begin{table}[h]
\centering
\small
\begin{tabular}{l|r|r|r}
\hline
Layer             & P-Emb                    & P-Sent & P-LM   \\ \hline
Embedding         & 300                      & 300    & 400     \\ \hline
Embedding noise   & 0.1                      & 0.1    & 0.1     \\ \hline
Embedding dropout & 0.2                      & 0.2    & 0.2     \\ \hline
LSTM size         & \multicolumn{1}{r|}{400} & 400    & 600/800 \\ \hline
LSTM dropout      & \multicolumn{1}{r|}{0.4} & 0.4    & 0.4     \\ \hline
\end{tabular}
\caption{Hyper-parameters of our models.}
\label{table:hparams}
\end{table}

\subsection{Official Results}\label{sec:results} 
Our team ranked \nth{3} out of 30 participants, achieving $ 0.703$ F1-score
on the official test set. Table \ref{table:iest} shows the official ranking of the top scoring teams.
\begin{table}[h]
\small
\captionsetup{farskip=0pt}
\centering
\begin{tabular}{c|c|c}
\hline
Rank       & Team Name         & Macro F1       \\ \hline
1          & Amobee            & 0.714          \\ \hline
2          & IIIDYT            & 0.710          \\ \hline
\textbf{3} & \textbf{NTUA-SLP} & \textbf{0.703} \\ \hline
4          & UBC-NLP           & 0.693          \\ \hline
5          & Sentylic          & 0.692    \\ \hline    
\end{tabular}

\caption{Results of the WASSA IEST competition.}
\label{table:iest}
\end{table}

\subsection{Experiments}\label{sec:experiments}
\noindent\textbf{Baselines}\label{sec:baselines}. 
In Table~\ref{table:finally} we compare the proposed TL approaches against two strong baselines: (1) a Bag-of-Words (BoW) model with TF-IDF weighting and (2) a Bag-of-Embeddings (BoE) model, where we retrieve the \textit{word2vec} representations of the words in a tweet and compute the tweet representation as the centroid of the constituent \textit{word2vec} representations.
Both \textit{BoW} and \textit{BoE} features are then fed to a linear SVM classifier, with tuned $C=0.6$.
All of our reported F1-scores are calculated on the evaluation (\textit{dev}) set, due to time constraints.

\noindent\textbf{\textit{P-Emb} and\textit{ P-Sent} models (\ref{sec:system1}, \ref{sec:system2})}\label{sec:embsent}. 
We evaluate the \textit{P-Emb} and \textit{P-Sent} models, using both bidirectional and unidirectional LSTMs.
The F1 score of our best models is shown in Table \ref{table:finally}. As expected, bi-LSTM models achieve higher performance.

\begin{table}[h!]
\centering
\small
\resizebox{\columnwidth}{!}{%
\begin{tabular}{c|ccc|r} \hline
LM Fine-tuning & \multicolumn{3}{c|}{LM Transfer}   &       \\ \hline
                        & Simple FT &  SGU & Concat. & F1    \\ \hline
\multirow{4}{*}{Simple FT}   &  \cmark         &             &         & 0.672 \\
                             &  \cmark         &             & \cmark        & 0.667 \\
                             &           &  \cmark           &         & 0.676 \\
                             &           &  \cmark           &  \cmark       & 0.673 \\ \hline
\multirow{4}{*}{ SGU} &  \cmark         &             &         & 0.673 \\
                             &  \cmark         &             &  \cmark       & 0.667 \\
                             &           &   \cmark          &         & 0.678 \\
                             &           &   \cmark          &  \cmark       &\textbf{ 0.682} \\ \hline
\end{tabular}
}
\caption{Results of the P-LM, trained on the EmoCorpus. The first column refers to the way we fine-tune each LM on the IEST dataset and the second to the way we finally fine-tune the classifier on the same dataset.}
\label{table:lms}

\end{table}

\begin{table}[h]
\centering
\small
\begin{tabular}{l|r}
\hline
Dataset    & F1 \\ \hline
EmoCorpus  & 0.682    \\ \hline
EmoCorpus+ & 0.680    \\ \hline
GenCorpus  & 0.675    \\ \hline
\end{tabular}
\caption{Comparison of the P-LM models, all fine-tuned with \textit{SGU} and \textit{Concat.} methods.}
\label{table:lm-comp}
\end{table}

\begin{table}[t]
\centering
\small
\begin{tabular}{l|r}
\hline
Model               & F1 \\ \hline
Bag of Words (BoW)        & 0.601    \\ \hline
Bag of Embeddings (BoE)   & 0.605    \\ \hline  \hhline{=|=}
P-Emb               & 0.668    \\ \hline
P-Sent             & 0.671    \\ \hline
P-LM                 & \textbf{0.675}    \\ \hline  \hhline{=|=}
P-Emb + bidir.      & 0.684    \\ \hline
P-Sent + bidir.     & 0.674    \\ \hline
P-LM + SGU           & 0.679    \\ \hline
P-LM + SGU + Concat. & 0.682    \\ \hline   \hhline{=|=}
Ensembling (UA) P-Emb + P-Sent  & 0.684    \\ \hline
Ensembling (UA) P-Sent + P-LM  & 0.695    \\ \hline
Ensembling (UA) P-Emb + P-LM  & 0.701    \\ \hline
\hhline{=|=}
Ensembling (MV) All    & 0.700    \\ \hline
Ensembling (UA) All    & \textbf{0.702}    \\ \hline
\end{tabular}
\caption{Results of our experiments when tested on the evaluation (\textit{dev}) set. \textit{BoW} and \textit{BoE} are our baselines, while \textit{P-Emb}, \textit{P-Sent} and \textit{P-LM} our proposed TL approaches. \textit{SGU} stands for Simplified Gradual Unfreezing, \textit{bidir.} for bi-LSTM, \textit{Concat.} for the concatenation method, \textit{UA} for Unweighted Average and \textit{MV} for Majority Voting ensembling.}
\label{table:finally}

\end{table}

\noindent\textbf{\textit{P-LM }(\ref{sec:system3})}\label{sec:plm}. 
For the experiments with the pretrained LMs, we intend to transfer not just the first layer of our network, but rather the whole model, so as to capture more high-level features of language. As mentioned above, there are three distinct steps concerning the training procedure of this TL approach: (1) \textit{LM pretraining}: we train three LMs on the EmoCorpus, EmoCorpus+ and GenCorpus corpora, (2) \textit{LM fine-tuning}: we fine-tune the LMs on the IEST dataset, with 2 different ways. The first one is  simple fine-tuning, while the second one is our simplified gradual unfreezing (SGU) technique. 
(3) \textit{LM transfer}: We now have 6 LMs, fine-tuned on the IEST dataset. We transfer their weights to our final emotion classifier, we add attention to the LSTM layers and we experiment again with our 2 ways of fine-tuning and the \textit{concatenation method} proposed in Sec. \ref{sec:system3}.


In Table \ref{table:lms} we present all possible combinations of transferring the \textit{P-LM} to the IEST task. We observe that SGU consistently outperforms Simple Fine-Tuning (Simple FT). Due to the difficulty in running experiments for all possible combinations, we compare our best approach, namely \textit{SGU} + \textit{Concat}., with \textit{P-LM}s trained on our three unlabeled Twitter corpora, as depicted in Table \ref{table:lm-comp}. Even though EmoCorpus contains less training examples,\textit{ P-LM}s trained on it learn to encode more useful information for the task at hand.

\noindent
\subsection{Ensembling}
\noindent Our submitted model is an ensemble of the models with the best performance. More specifically, we leverage the following models: (1) TL of pretrained word embeddings, (2) TL of pretrained sentiment classifier, (3) TL of 3 different LMs, trained on 2M, 4M and 5M respectively. We use Unweighted Average (UA) ensembling  of our best models from all aforementioned approaches. Our final results on the evaluation data are shown in Table \ref{table:finally}.

\subsection{Discussion}
	As shown in Table \ref{table:finally}, we observe that all of our proposed models achieve individually better performance than our baselines by a large margin.
    Moreover, we notice that, when the three models are trained with unidirectional LSTM and the same number of parameters, the \textit{P-LM} outperforms both the \textit{P-Emb} and the \textit{P-Sent} models.
    As expected, the upgrade to bi-LSTM improves the results of \textit{P-Emb} and \textit{P-Sent}. 
    We hypothesize that \textit{P-LM} with bidirectional pretrained language models would have outperformed both of them. 
    Furthermore, we conclude that both SGU for fine-tuning and the concatenation method enhance the performance of the\textit{ \textit{P-LM} }approach. 
  As far as the ensembling is concerned, both approaches, \textit{MV} and \textit{UA}, yield similar performance improvement over the individual models.
  In particular, we notice that adding the \textit{P-LM} predictions to the ensemble contributes the most.
  This indicates that \textit{P-LMs} encode more diverse information compared to the other approaches.

\section{Conclusion}
In this paper we describe our deep-learning methods for missing emotion words classification, in the Twitter domain. We achieved very competitive results in the IEST competition, ranking 3$^{rd}$/30 teams.
The proposed approach is based on an
ensemble of Transfer Learning techniques. We demonstrate that the use of refined, high-level features of text, as the ones encoded in language models, yields a higher performance. In the future, we aim to experiment with subword-level models, as they have shown to consistently face the OOV words problem ~\cite{sennrich2015neural,bojanowski2016enriching}, which is more evident in Twitter. Moreover, we would like to explore other transfer learning approaches.

Finally, we share the source code of our models~\footnote{\url{/github.com/alexandra-chron/wassa-2018}}, in order to make our results reproducible and facilitate further experimentation in the field.

\bibliography{refs}
\bibliographystyle{acl_natbib}

\appendix

\end{document}